\documentclass[twoside]{article}
\usepackage[accepted]{aistats2012}
\usepackage{graphicx}
\usepackage{subfig}
\usepackage{color}
\usepackage{url}
\usepackage{amsmath}
\usepackage{amssymb}
\usepackage[mathscr]{eucal}
\usepackage{macros}
\usepackage[round]{natbib}
\usepackage{multirow}
\usepackage{booktabs}
\usepackage{caption}
\usepackage{rotating}

\makeatletter
\let\@copyrightspace\relax
\makeatother
\begin{document}

\twocolumn[

\runningtitle{On Nonparametric Guidance for Learning Autoencoder Representations}
\aistatstitle{On Nonparametric Guidance for \\Learning Autoencoder Representations}

\aistatsauthor{Jasper Snoek \And Ryan Prescott Adams \And Hugo Larochelle}

\aistatsaddress{ University of Toronto \And Harvard University \And
  University of Sherbrooke } ]

\begin{abstract}
  Unsupervised discovery of latent representations, in addition to
  being useful for density modeling, visualisation and exploratory
  data analysis, is also increasingly important for learning features
  relevant to discriminative tasks.  Autoencoders, in particular, have
  proven to be an effective way to learn latent codes that reflect
  meaningful variations in data.  A continuing challenge, however, is
  guiding an autoencoder toward representations that are useful for
  particular tasks.  A complementary challenge is to find codes that
  are invariant to irrelevant transformations of the data.  The most
  common way of introducing such problem-specific guidance in
  autoencoders has been through the incorporation of a parametric
  component that ties the latent representation to the label
  information.  In this work, we argue that a preferable approach
  relies instead on a nonparametric guidance mechanism.  Conceptually,
  it ensures that there exists a function that can predict the label
  information, without explicitly instantiating that function. The
  superiority of this guidance mechanism is confirmed on two
  datasets. In particular, this approach is able to incorporate
  invariance information (lighting, elevation, etc.) from the small
  NORB object recognition dataset and yields state-of-the-art
  performance for a single layer, non-convolutional network.
\end{abstract}

\section{Introduction}
The inference of constrained latent representations plays a key role
in machine learning and probabilistic modeling.  Broadly, the idea is
that discovering a \emph{compressed} representation of the data will
correspond to determining what is important and unimportant about the
data.  One can also view constrained latent representations as
providing \textit{features} that can be used to solve other machine
learning tasks.  Of particular importance are methods for latent
representation that can efficiently construct codes for out-of-sample
data, enabling rapid feature extraction.  Neural networks, for
example, provide such feed forward feature extractors, and
autoencoders, specifically, have found use in domains such as image
classification~\citep{VincentP2008}, speech recognition
\citep{deng-etal-2010a} and Bayesian nonparametric models
\citep{adams-etal-2010a}.

While the representations learned with autoencoders are often useful
for discriminative tasks, they require that the salient variations in
the data distribution be relevant for labeling.  This is not
necessarily always the case; as irrelevant factors of variation grow
in importance and increasingly dominate the input distribution, the
representation extracted by autoencoders tends to become less
useful~\citep{LarochelleH2007}. To address this issue,
\citet{BengioY2006} introduced mild supervised guidance into the
autoencoder training objective, by adding connections from the hidden
layer to output units predicting label information (those connections
are equivalent to the parameters of a logistic regression
classifier). The same approach was followed by \citet{RanzatoM2008},
to learn compact representations of documents.

One downside of this approach is that it potentially complicates the
task of learning the autoencoder representation. Indeed, it now tries
to solve two additional problems: find a hidden representation from
which the label information can be predicted \textit{and} track the
parametric value of that predictor (i.e.\ the logistic regression
weights) throughout learning. However, we are only interested in the
first problem (increased predictability of the label). The actual
parametric value of the label predictor is not important. Once the
autoencoder is trained, the label predictor can easily be found by
training a logistic regressor from scratch, keeping the hidden layer
fixed. We might even want to use a classifier that is very different
from the logistic regression classifier for which the hidden layer has been
trained for.

In this work, we investigate this issue and explore a different
approach to introducing supervised guidance. We treat the latent space
of the autoencoder as also being the latent space for a Gaussian
process latent variable model (GPLVM) \citep{lawrence-2005a}.  The
discriminative labels are then taken to belong to the visible space of
the GPLVM. The end result is a nonparametrically guided autoencoder which
combines an efficient feed-forward parametric encoder/decoder with the
Bayesian nonparametric inclusion of label information. We discuss how
this corresponds to marginalizing out the parameters of a mapping from
the latent representation to the label and show experimentally how
this approach is preferable to explicitly instantiating such a
parametric mapping.  Finally, we show how this hybrid model also
provides a way to guide the autoencoder's representation \textit{away}
from irrelevant features to which the encoding should be invariant.

\section{Unsupervised Learning of Latent Representations}

We first review the two different latent representation learning algorithms on
which this work builds. We then discuss a relationship between the two
that provides part of the motivation for the proposed nonparametrically
guided autoencoder.

\subsection{Autoencoder Neural Networks}
Our starting point is the
autoencoder \citep{Cottrell87}, which is an artificial neural network
that is trained to reproduce (or reconstruct) the input at its
output. Its computations are decomposed into two parts: the encoder,
which computes a latent (often lower-dimensional) representation of
the input, and the decoder, which reconstructs the original input from
its latent representation.  We denote the latent space by~$\mcX$ and
the visible (data) space by~$\mcY$.  We assume that these spaces are
real-valued with dimension~$J$ and~$K$, respectively,
i.e.,~${\mcX\medeq\reals^J}$ and~${\mcY\medeq\reals^K}$.  We denote
the encoder, then, as a function~${g(\boldy\,;\,\phi):\mcY\to\mcX}$
and the decoder as~${f(\boldx\,;\,\psi):\mcX\to\mcY}$.  With a set
of~$N$
examples~${\mcD\medeq\{\boldy^{(n)}\}^N_{n=1},\,\boldy^{(n)}\in\mcY}$,
we jointly optimize the encoder parameters~$\phi$ and decoder
parameters~$\psi$ for the least-squares reconstruction cost:
\begin{multline}
  \phi^\star,\psi^\star
  \medeq \arg\min_{\phi,\psi} \sum_{n=1}^N \sum^K_{k=1} (y^{(n)}_k -
  f_k(g(\boldy^{(n)}; \phi);\psi))^2,
\end{multline}
where~$f_k(\cdot)$ is the~$k$th output dimension of~$f(\cdot)$.
Autoencoders have become popular as a module for ``greedy
pre-training'' of deep neural networks~\citep{BengioY2006}.  In
particular, the \textit{denoising} autoencoder of \citet{VincentP2008}
is effective at learning overcomplete latent representations,~i.e.,
codes of higher dimensionality than the input.  Overcomplete
representations are thought to be ideal for discriminative tasks, but
are difficult to learn due to trivial ``identity'' solutions to the
autoencoder objective.  This problem is circumvented in the denoising
autoencoder by providing as input a corrupted training example, while
evaluating reconstruction on the noiseless original.  With this
objective, the autoencoder learns to leverage the statistical
structure of the inputs to extract a richer latent representation.

\subsection{Gaussian Process Latent Variable Models}
One alternative approach to the learning of latent representations is
to consider a lower-dimensional \textit{manifold} that reflects the
statistical structure of the data.  Such manifolds may be difficult to
directly define, however, and so many approaches to latent coding
frame the problem indirectly by specifying distributions on functions
between the visible and latent spaces.  The Gaussian process latent
variable model (GPLVM) of \citet{lawrence-2005a} takes a Bayesian
probabilistic approach to this and constructs a distribution over
mapping functions using a Gaussian process (GP) prior.  The GPLVM
results in a powerful nonparametric model that analytically
marginalizes over the infinite number of possible mappings from the
latent to the visible space. While initially used for visualization of
high dimensional data, GPLVMs have achieved state-of-the-art results
for a number of tasks, including modeling human motion
\citep{wang-etal-2008a}, classification \citep{urtasun-darrell-2007a}
and collaborative filtering \citep{lawrence-urtasun-2009a}.

As in the autoencoder, the GPLVM assumes that the~$N$ observed data
${\mcD\medeq\{\boldy^{(n)}\}^N_{n=1}}$ are the image of a homologous
set~${\{\boldx^{(n)}\}^N_{n=1}}$, arising from a vector-valued
``decoder'' function ${f(\boldx):\mcX\to\mcY}$.  Analogously to the
squared-loss of the previous section, the GPLVM assumes that the
observed data have been corrupted by zero-mean Gaussian
noise:~${\boldy^{(n)}\!=\!f(\boldx^{(n)})\!+\!\varepsilon}$
with~${\varepsilon\!\sim\!\distNorm(0, \sigma^2 \bbI_{K})}$.  The
innovation of the GPLVM is to place a Gaussian process prior on the
function~$f(\boldx)$ and then optimize the latent
representation~$\{\boldx^{(n)}\}^N_{n=1}$, while marginalizing out the
unknown~$f(\boldx)$.

\subsubsection{Gaussian Process Priors}
The Gaussian process provides a flexible distribution over random
functions, the properties of which can be specified via a positive
definite covariance function, without having to choose a particular
finite basis.  Typically, Gaussian processes are defined in terms of a
distribution over scalar functions and in keeping with the convention
for the GPLVM, we shall assume that~$K$ independent GPs are used to
construct the vector-valued function~$f(\boldx)$.  We denote each of
these functions as~${f_k(x):\mcX\to\reals}$.  The GP requires a
covariance kernel function, which we denote
as~${C(\boldx,\boldx'):\mcX\stimes\mcX\to\reals}$.  The defining
characteristic of the GP is that for any finite set of~$N$ data
in~$\mcX$ there is a corresponding~$N$-dimensional Gaussian
distribution over the function values, which in the GPLVM we take to
be the components of~$\mcY$.  The~${N\stimes N}$ covariance matrix of
this distribution is the matrix arising from the application of the
covariance kernel to the~$N$ points in~$\mcX$.  We denote any
additional parameters governing the behavior of the covariance
function by~$\theta$.

Under the component-wise independence assumptions of the GPLVM, the
Gaussian process prior allows one to analytically integrate out
the~$K$ latent scalar functions from~$\mcX$ to~$\mcY$.  Allowing for
each of the~$K$ Gaussian processes to have unique
hyperparameter~$\theta_k$, we write the marginal likelihood, i.e., the
probability of the observed data given the hyperparameters and the
latent representation, as
\begin{multline}
  \label{eqn:gplvm-ml}
  p(\{\boldy^{(n)}\}^N_{n=1}\given\{\boldx^{(n)}\}^N_{n=1},\{\theta_k\}^K_{k=1},\sigma^2) \\
  = \prod^K_{k=1}\distNorm(\boldy_k^{(\cdot)}\given 0,
  \boldSigma_{\theta_k} \splus \sigma^2\bbI_{N}),
\end{multline}
where $\boldy_k^{(\cdot)}$ refers to the vector
$[y_k^{(1)},\dots,y_k^{(N)}]$ and where~$\boldSigma_{\theta_k}$ is the
matrix arising from~$\{\boldx_n\}^N_{n=1}$ and~$\theta_k$.  In the
basic GPLVM, the optimal~$\boldx_n$ are found by maximizing this
marginal likelihood.

\subsubsection{The Back-Constrained GPLVM}
Although the GPLVM enforces a smooth mapping from the latent
representation to the observed data, the converse is not true:
neighbors in observed space need not be neighbors in the latent
representation.  In many applications this can be an undesirable
property.  Furthermore, encoding novel datapoints into the latent
space is not straightforward in the GPLVM; one must optimize the
latent representations of out-of-sample data using, e.g., conjugate
gradient methods.  With these considerations in mind,
\citet{lawrence-candela-2006a} reformulated the GPLVM with the
constraint that the hidden representation be the result of a smooth
map from the observed space.  Parameterized by~$\phi$, this
``encoder'' function is denoted as~${g(\boldy\,;\,\phi):\mcY\to\mcX}$.
The marginal likelihood objective of this \textit{back-constrained}
GPLVM can now be formulated as finding the optimal~$\phi$ under:
\begin{multline}
  \label{eqn:gplvm-back-obj}
  \phi^\star \medeq
  \arg\min_\phi\sum^K_{k=1}\ln|\boldSigma_{\theta_k,\phi}\splus\sigma^2\bbI_{N}|
  \\ +
  {\boldy_k^{(\cdot)}}^\trans(\boldSigma_{\theta_k,\phi}\splus\sigma^2\bbI_{N})^{-1}
  \boldy_k^{(\cdot)},
\end{multline}
where the~$k$th covariance matrix~$\boldSigma_{\theta_k,\phi}$ now
depends not only on the kernel hyperparameters~$\theta_k$, but also on
the parameters of~$g(\boldy\,;\,\phi)$, i.e.,
\begin{align}
  \label{eqn:gplvm-back-cov}
[\boldSigma_{\theta_k,\phi}]_{n,n'} &=
C( g(\boldy^{(n)};\phi), g(\boldy^{(n')};\phi)\,;\,\theta_k ).
\end{align}
\citet{lawrence-candela-2006a} explored multilayer perceptrons and
radial-basis-function networks as possible smooth
maps~$g(\boldy\,;\,\phi)$.

\subsection{GPLVM as an Infinite Autoencoder}
The relationship between Gaussian processes and artificial neural
networks is well-established.  \citet{neal-96} showed that the prior
over functions implied by many parametric neural networks becomes a GP
in the limit of an infinite number of hidden units, and
\citet{williams-1998a} subsequently derived a covariance function that
corresponds to such a network under a particular activation function.

One overlooked consequence of this relationship is that it also
connects autoencoders and the back-constrained Gaussian process latent
variable model.  By applying the covariance function of
\citet{williams-1998a} to the GPLVM, the resulting model is a density
network \citep{mackay-94} with an infinite number of hidden units in
the single hidden layer.  Then, using a neural network for the GPLVM
backconstraints transforms the density network into a semiparametric
autoencoder, where the encoder is a parametric neural network and the
decoder is a Gaussian process.

Alternatively, one can start from the autoencoder and notice that, for
a linear decoder with a least-squares reconstruction cost and
zero-mean Gaussian prior over its weights, it is possible to integrate
out the decoder.  Learning then corresponds to the minimization of
Eqn.~(\ref{eqn:gplvm-back-obj}) with a linear kernel for
Eqn.~(\ref{eqn:gplvm-back-cov}).  Any non-degenerate positive definite
kernel corresponds to a decoder of infinite size, and also recovers
the general back-constrained GPLVM algorithm.
 
Such an infinite autoencoder exhibits some desirable properties.  The
infinite decoder network obviates the need to explicitly specify and
learn a parametric form for the generally superfluous decoder network
and rather marginalises over all possible decoders. This comes at the
cost of having to invert as many matrices (the GP covariances) as
there are input dimensions. Hence, for large input dimensionality, one
could argue that the fully parametric autoencoder is preferable.

\section{Supervised Guiding of Latent Representations}
\label{sec:sglr}
As discussed earlier, when the salient variations in the input are
only weakly informative about a particular discriminative task, it can
be useful to incorporate label information into unsupervised learning.
\citet{BengioY2006} showed, for example, that while a purely
supervised signal can lead to overfitting, mild supervised guidance
can be beneficial when initializing a discriminative deep neural
network.  For that reason, \citet{BengioY2006} proposed that latent
representations also be trained to predict the label information, by
adding a parametric mapping~${c(\boldx\,;\,\Lambda):\mcX\to\mcZ}$ from
the latent representation's space~$\mcX$ to the label space~$\mcZ$ and
backpropagating error gradients from the output to the representation.
\citet{BengioY2006} investigated the use of a linear logistic
regression classifier for the parametric mapping. Such ``partial
supervision'' would encourage discovery of a latent representation
that is useful to a specific (but learned) parametrization of such a
linear classifier.  A similar approach was used by
\citet{RanzatoM2008} to learn compact representations of documents.

There are two disadvantages to this strategy.  First, the assumption
of a specific parametric form for the mapping~$c(\boldx\,;\,\Lambda)$
restricts the guidance to classifiers within that family of mappings.
The second is that the learned representation is committed to one
particular setting of the parameters~$\Lambda$.  Consider the learning
dynamics of gradient descent optimization for this strategy.  At every
iteration $t$ of descent (with current state
$\phi_t,\psi_t,\Lambda_t$), the gradient from supervised guidance
encourages the latent representation (currently parametrized by
$\phi_t,\psi_t$) to become more predictive of the labels under the
current label map $c(\boldx\,;\,\Lambda_t)$.  Such behavior
discourages moves in $\phi,\psi$ space that make the latent
representation more predictive under some other label map
$c(\boldx\,;\,\Lambda^\star)$ where $\Lambda^\star$ is potentially
distant from $\Lambda_t$. Hence, while the problem would seem to be
alleviated by the fact that~$\Lambda$ is learned jointly, this
constant pressure towards representations that are immediately useful
should increase the difficulty of representation learning.

\subsection{Nonparametrically Guided Autoencoder}
Rather than directly specifying a particular discriminative regressor
for guiding the latent representation, it seems more desirable
to simply ensure that such a function \textit{exists}.  That is, we
would prefer not to have to choose a latent representation that is
tied to a specific map to labels, but instead find representations
that are consistent with many such maps.  One way to arrive at such a
guidance mechanism is to marginalize out the parameters~$\Lambda$ of a
label map~$c(\boldx\,;\,\Lambda)$ under a distribution that permits a
wide family of functions.  We have seen previously that this can be
done for reconstructions of the input space with a
decoder~$f(\boldx\,;\,\psi)$.  We follow the same reasoning and do
this instead for~$c(\boldx\,;\,\Lambda)$.  Integrating out the
parameters of the label map yields a back-constrained GPLVM acting on
the label space~$\mcZ$, where the back constraints are determined by
the input space~$\mcY$.  The positive definite kernel specifying the
Gaussian process then determines the properties of the distribution
over mappings from the latent representation to the labels.  The
result is a hybrid of the autoencoder and back-constrained GPLVM,
where the encoder is shared across models.  For notation, we will
refer to this approach to guided latent representation as a
\textit{nonparametrically guided autoencoder}, or NPGA.

Let the label space~$\mcZ$ be an~$M$-dimensional real
space\footnote{For discrete labels, we use a ``one-hot'' encoding.},
i.e.,~${\mcZ\!=\!\reals^M}$, and the~$n$th training example has a
label vector~$\boldz^{(n)}\in\mcZ$.  The covariance function that
relates label vectors in the NPGA is
\begin{align*}
[\boldSigma_{\theta_m,\phi,\bGamma}]_{n,n'} &=
C( \bGamma\cdot g(\boldy^{(n)};\phi), \bGamma\cdot g(\boldy^{(n')};\phi)\,;\,\theta_m ),
\end{align*}
where~${\bGamma \in \reals^{H \stimes J}}$ is an $H$-dimensional linear
projection of the encoder output.  For ${H\ll J}$, this projection
improves efficiency and reduces overfitting. Learning in the NPGA is
then formulated as finding the optimal~$\phi,\psi,\bGamma$ under the
combined objective:
\begin{align*}
  \phi^\star,\psi^\star,\bGamma^\star &\medeq \arg\min_{\phi,\psi,\bGamma} (1\sminus\alpha) L_{\rm auto}(\phi,\psi) + \alpha L_{\rm GP}(\phi,\bGamma)
\end{align*}
where $\alpha\in[0,1]$ linearly blends the two objectives
\begin{align*}
L_{\rm auto}(\phi,\psi) &= \frac{1}{K} \sum_{n=1}^N\sum^K_{k=1} (y^{(n)}_k - f_k(g(\boldy^{(n)}; \phi);\psi))^2 \\
L_{\rm GP}(\phi,\bGamma) &= \frac{1}{M}
\sum^M_{m=1}\left[\vphantom{{\boldz_m^{(\cdot)}}^\trans}
  \ln|\boldSigma_{\theta_m,\phi,\bGamma}\splus\sigma^2\bbI_{N}| \right. \\ 
&\qquad \left. + {\boldz_m^{(\cdot)}}^\trans(\boldSigma_{\theta_m,\phi,\bGamma}\splus\sigma^2\bbI_{N})^{-1} \boldz_m^{(\cdot)} \right].
\end{align*}

We use a linear decoder for $f(\boldx\,;\,\psi)$, and the encoder
$g(\boldy;\phi)$ is a linear transformation followed by a fixed
element-wise nonlinearity.  

As is common for
autoencoders and to reduce the number of free parameters in the model,
the encoder and decoder weights are tied. For the larger NORB dataset, we
divide the training data into mini-batches of 350 training cases and
perform three iterations of conjugate gradient descent per
mini-batch. Finally, as proposed in the denoising autoencoder variant
of \citet{VincentP2008}, we always add noise to the
encoder inputs in cost $L_{\rm auto}(\phi,\psi)$, keeping the noise
fixed during each iteration.

\subsection{Related Models}
The combination of parametric unsupervised learning and nonparametric
supervised learning has been examined previously.
\citet{salakhutdinov-hinton-2007a} proposed merging autoencoder
training with nonlinear neighborhood component analysis, which
encourages the encoder to have similar outputs for similar inputs
belonging to the same class. Note that the backconstrained-GPLVM
performs a similar role. Examining Equation \ref{eqn:gplvm-back-obj},
one can see that the first term, the log determinant of the kernel,
regularizes the latent space. It pulls all examples together as the
determinant is minimized when the covariance between all pairs is
maximized. The second term is a data fit term, pushing examples that
are distant in label space apart in the latent space.  In the case of
a one-hot coding, the labels act as indicator variables including only
indices of the concentration matrix that reflect inter-class pairs in
the loss. Thus the GPLVM enforces that examples close in the label
space will be closer in the latent space than examples that are
distant in label space. There are several notable differences,
however, between this work and the NPGA. First, as the NPGA is a
natural generalization of the back-constrained GPLVM, it can be
intuitively interpreted as a marginalization of label maps, as
discussed in the previous section. Second, the NPGA enables the wide
library of covariance functions from the Gaussian process literature
to be incorporated into the framework of learning guided
representation and naturally accomodates continuous labels. Finally,
as will be discussed in Section~\ref{sec:norb}, the NPGA not only
enables learning of unsupervised features that capture
discriminatively-relevant information, but also allows representations
that can \textit{ignore} irrelevant information.

Previous work has also hybridized Gaussian processes and unsupervised
connectionist learning.  In \citet{salakhutdinov-hinton-2008a},
restricted Boltzmann machines were used to initialize a neural network
that would provide features to a Gaussian process regressor or
classifier.  Unlike the NPGA, however, this approach does not address
the issue of guided unsupervised representation. Indeed, in NPGA,
Gaussian processes are used only for representation learning, are
applied only on small mini-batches and are not required at test time.
This is important, since deploying a Gaussian process on large
datasets such as the NORB data poses significant practical
problems. Because their method relies on a Gaussian process at test
time, a direct application of the approach proposed by
\citet{salakhutdinov-hinton-2008a} would be prohibitively slow.

Although the GPLVM was originally proposed as a latent variable model
conditioned on the data, there has been work on adding discriminative
label information and additional signals.  The Discriminative GPLVM
(DGPLVM) \citep{urtasun-darrell-2007a} incorporates discriminative
class labels through a prior based on discriminant analysis
that enforces separability between classes in the latent space.
The DGPLVM is, however, restricted to discrete labels and requires a
GP mapping to the data, which is computationally prohibitive for
high dimensional data.  \citet{shon-etal-2005a} introduced
a Shared-GPLVM (SGPLVM) that used multiple GPs to map from a single
shared latent space to various related signals.
\citet{wang-etal-2007a} demonstrate that a generalisation of
multilinear models arises as a GPLVM with product kernels, each
mapping to different signals. This allows one to separate various
signals in the data within the context of the GPLVM.  Again, due to
the Gaussian process mapping to the data, the shared and multifactor
GPLVM are not feasible on high dimensional data.  Our model overcomes the
limitations of these through using a natural parametric form of the
GPLVM, the autoencoder, to map to the data.

\section{Empirical Analyses}
\label{sec:experiments}

We now present experiments with NPGA on two different classification
datasets. Our implementation of NPGA is available for download at
\url{http://removed.for.anonymity.org}. In all experiments, the
discriminative value of the learned representation is evaluated by
training a linear (logistic) classifier, a standard practice for
evaluating latent representations.

\begin{figure*}[ht]
\vskip -0.1cm%
  \centering%
  \subfloat[\label{fig:oil_results_a}]{%
    \raisebox{0.05cm}{
    \includegraphics[width=0.25\textwidth]%
    {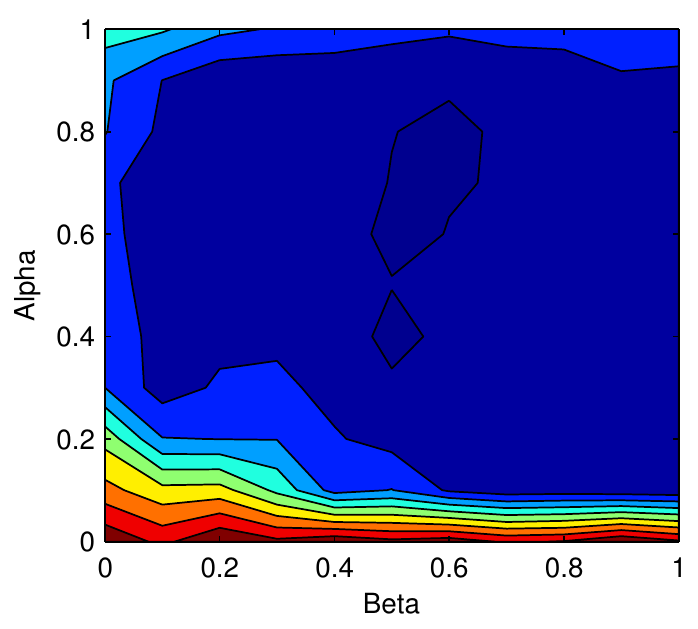}}}\quad%
  \subfloat[\label{fig:oil_results_b}]{%
    \includegraphics[width=0.245\textwidth]%
    {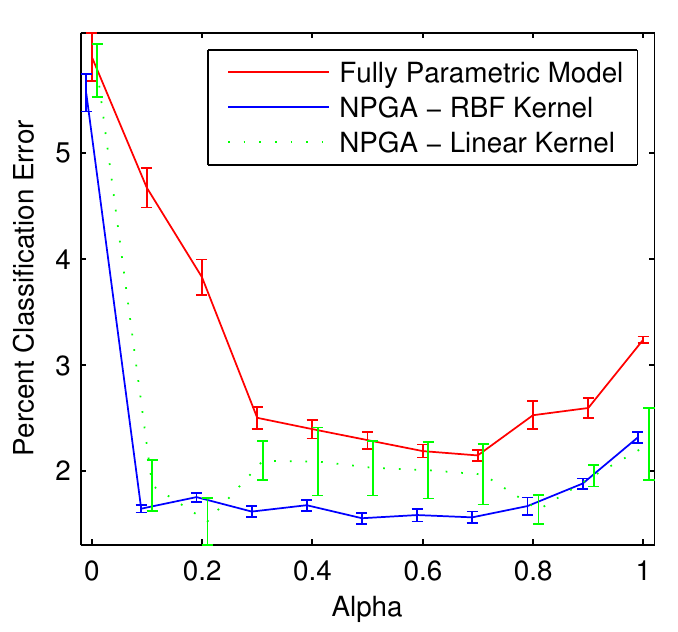}}\quad%
  \subfloat[\label{fig:oil_results_d}]{%
    \raisebox{0.4cm}{
    \includegraphics[width=0.21\textwidth]%
    {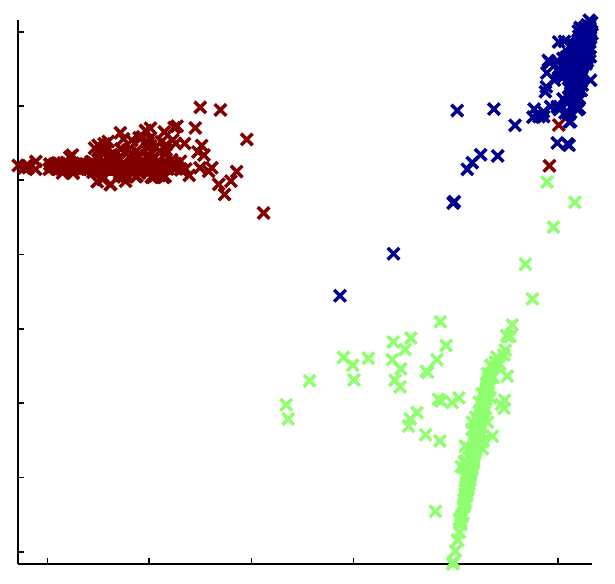}}}\quad%
  \subfloat[\label{fig:oil_results_e}]{%
    \raisebox{0.4cm}{
    \includegraphics[width=0.21\textwidth]%
    {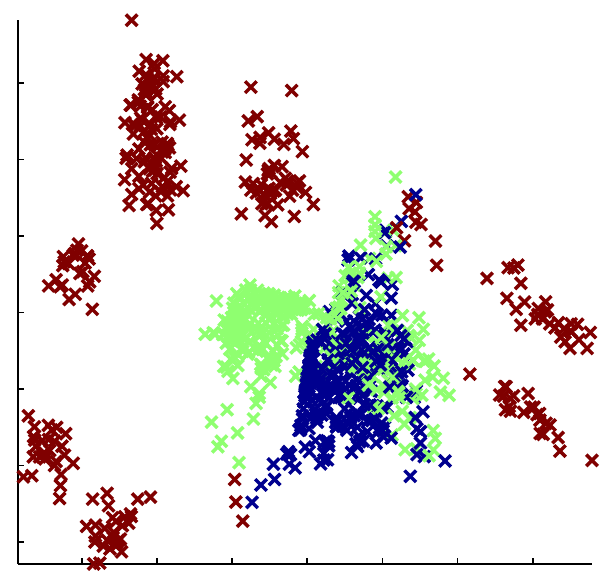}}}%
  \vskip -0.1cm%
  \caption{The effect of scaling the relative contributions of the
    autoencoder, logistic regressor and GP costs in the hybrid
    objective by modifying~$\alpha$ and~$\beta$.  (a)~Classification
    error on the test set on a linear scale from ~6\% (dark red) to
    ~1\% (dark blue) (b)~Cross-sections of (a) at~${\beta\!=\!0}$ (a
    fully parametric model) and~${\beta\!=\!1}$ (NPGA). % (c)~NPGA
    ($\alpha\medeq 0.5$)
    (c \& d)~Latent projections
    of the 1000 test cases within the latent space of the GP for a
    NPGA ($\alpha=0.5$) and a back-constrained GPLVM. }
\label{fig:oil_results}
\vspace{-0.5cm}
\end{figure*}

\subsection{Oil Flow Data}
As an initial empirical analysis we consider a multi-phase oil flow
classification problem \citep{bishop-james-1993a}.  The data are
twelve-dimensional, real-valued measurements of gamma densitometry
measurements from a simulation of multi-phase oil flow.  The
classification task is to determine from which of three phase
configurations each example originates.  There are 1,000 training and
1,000 test examples.  The relatively small size of these training data
make them useful for empirical evaluation of different models and
training procedures.  We use these data primarily to address two
concerns:
\begin{itemize}
\item %First, 
  To what extent does the nonparametric guidance of an
unsupervised parametric autoencoder improve the learned feature
representation with respect to the classification objective?
\item % Second,
What additional benefit is gained through using nonparametric guidance
over simply incorporating a parametric mapping to the labels?
\end{itemize}
To address these questions, we construct a new objective that linearly
blends our proposed supervised guidance cost $L_{\rm GP}(\phi,\bGamma)$
with the one proposed by \citet{BengioY2006},
referred to as $L_{\rm LR}(\phi,\bGamma)$:
\begin{align*}
  L(\phi,\psi,\Lambda,\bGamma\,;\,\alpha,\beta) &= (1\sminus\alpha)L_{\rm
    auto}(\phi,\psi) \\ &+ \alpha ((1\sminus\beta)L_{\rm LR}(\phi,\Lambda) 
  \\&+ \beta L_{\rm GP}(\phi,\bGamma)),
\end{align*}
where ${\beta \in [0,1]}$.  $\Lambda$ are the parameters of a
multi-class logistic regressor that maps to the labels.  Thus,
$\alpha$ controls the relative importance of supervised guidance,
while $\beta$ controls the relative importance of the parametric and
nonparametric supervised guidance.

A grid search over $\alpha$ and $\beta$ was performed at intervals
of~$0.1$ to assess the benefit of the nonparametric guidance.  At each
interval a model was trained for 100 iterations and classification
performance was assessed via logistic regression on the hidden units
of the encoder.  Notice how the cost~$L_{\rm LR}(\phi,\Lambda)$ is
specifically tailored to this situation.  The encoder used 250 noisy
rectified linear (NRenLU \citep{nair-hinton-2010a}) units, and
zero-mean Gaussian noise with a standard deviation of 0.05 was added
to the inputs of the autoencoder cost.  A subset of 100 training
samples was used to make the problem more challenging.  Each
experiment was repeated 20 times with random initializations.  The GP
label mapping used an RBF kernel and worked on a projected space of
dimension ${H\medeq 2}$.

Results are presented in Fig.~\ref{fig:oil_results}.
Fig.~\ref{fig:oil_results_b} demonstrates that performance improves
by integrating out the label map, even when compared with direct
optimization under the discriminative family that will be used at test
time.  Figs.~\ref{fig:oil_results_d}~and~\ref{fig:oil_results_e}
provide a visualisation of the latent representation learned by NPGA
and a standard back-constrained GPLVM.  We see that the former embeds
much more class-relevant structure than the latter.

An interesting observation is that a simple linear kernel also tends
to outperform parametric guidance (see Fig.~\ref{fig:oil_results_b}).
This doesn't mean that any kernel will work for any problem. However,
this confirms that the benefit of our approach is achieved mainly
through integrating out the label mapping, rather than having a more
powerful nonlinear mapping to the label.

\subsection{Small NORB Image Data}
\label{sec:norb}
As a second empirical analysis, the NPGA is evaluated on a
challenging dataset with multiple discrete and real-valued labels.
The small NORB data~\citep{lecun-etal-2004a} are stereo image pairs of
fifty toys belonging to five generic categories.  Each object was
imaged under six lighting conditions, nine elevations and eighteen
azimuths.  The objects were divided evenly into test and training sets
yielding 24,300 examples each.

The variations in the data resulting from the different
imaging conditions impose significant nuisance structure that will
invariably be learned by a standard autoencoder.  Fortunately, these
variations are known \emph{a priori}. In addition to the class labels, there
are two real-valued vectors (elevation and azimuth) and one discrete
vector (lighting type) associated with each image.  In our empirical
analysis we examine two questions:
\begin{itemize}
\item As the autoencoder attempts to
  coalesce the various sources of structure into its hidden layer, can
  the NPGA guide the learning in such a way as to separate the
  class-invariant transformations of the data from the class-relevant
  information?
\item Are the benefits of nonparametric guidance still observed in a
  larger scale classification problem, when mini-batch training is
  used?%
\end{itemize}%
\begin{figure}[t]%
\vspace{-0.575cm}%
  \centering%
  \newcommand{\norbfigsize}{0.176\textwidth}%
  \begin{sideways}~~~~~~Classes\end{sideways} \subfloat{%
    \includegraphics[width=0.176\textwidth]%
    {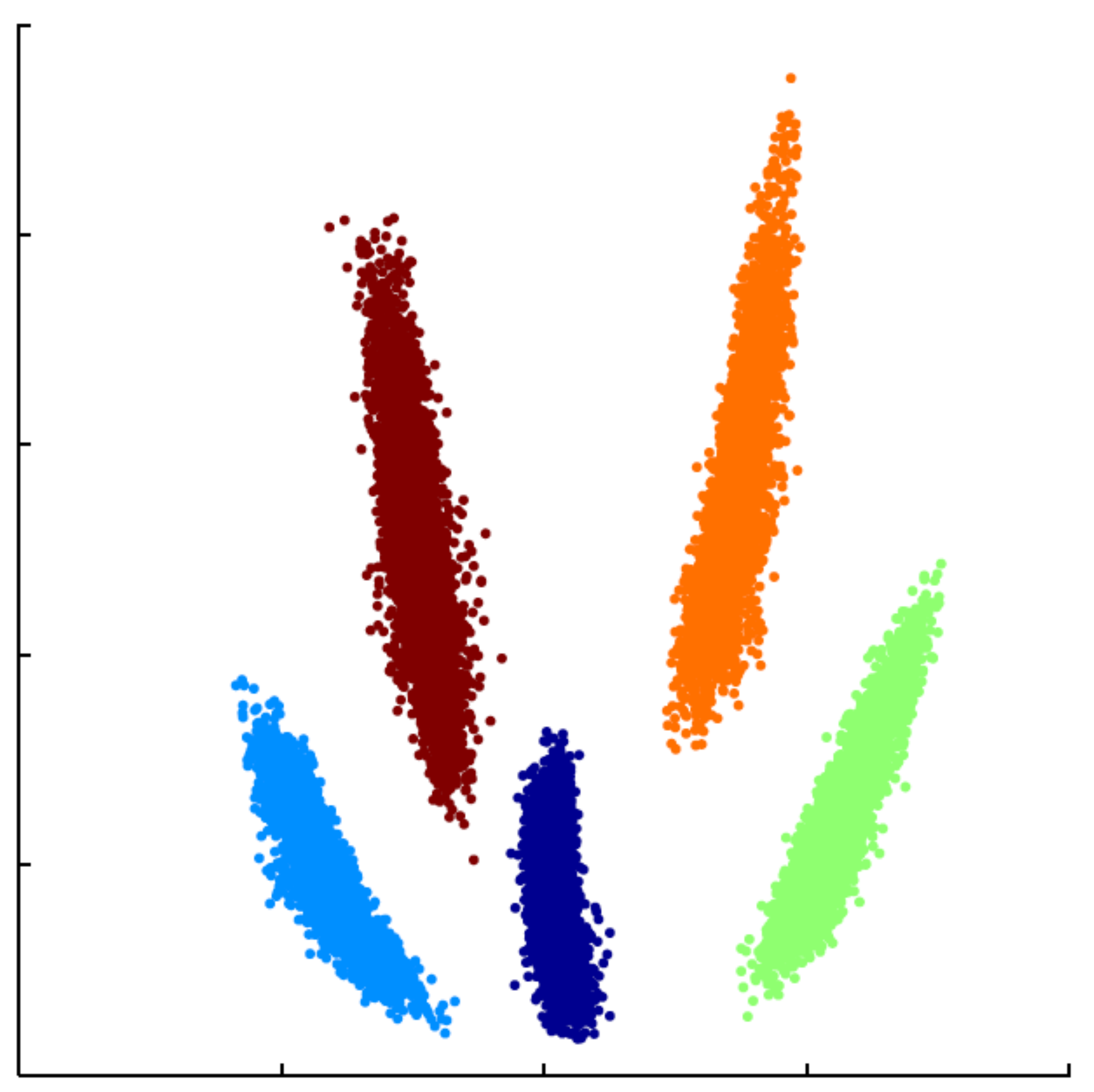}}\qquad%
  \subfloat{%
    \includegraphics[width=0.176\textwidth]%
    {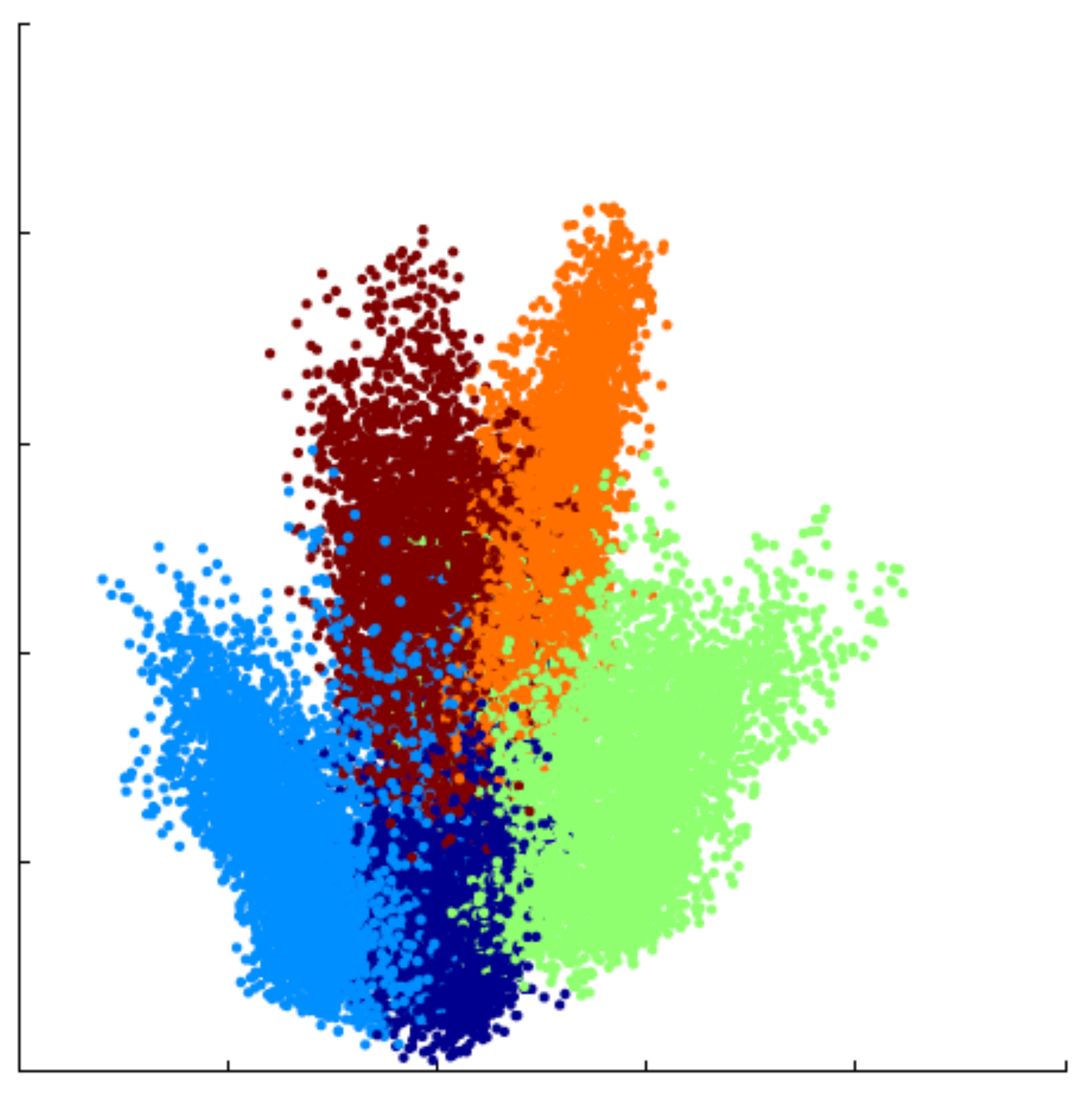}}\qquad%
  \\
  \vspace{-0.3cm}%
  \begin{sideways}~~~~~Elevation\end{sideways}
  \subfloat{%
    \includegraphics[width=\norbfigsize]%
    {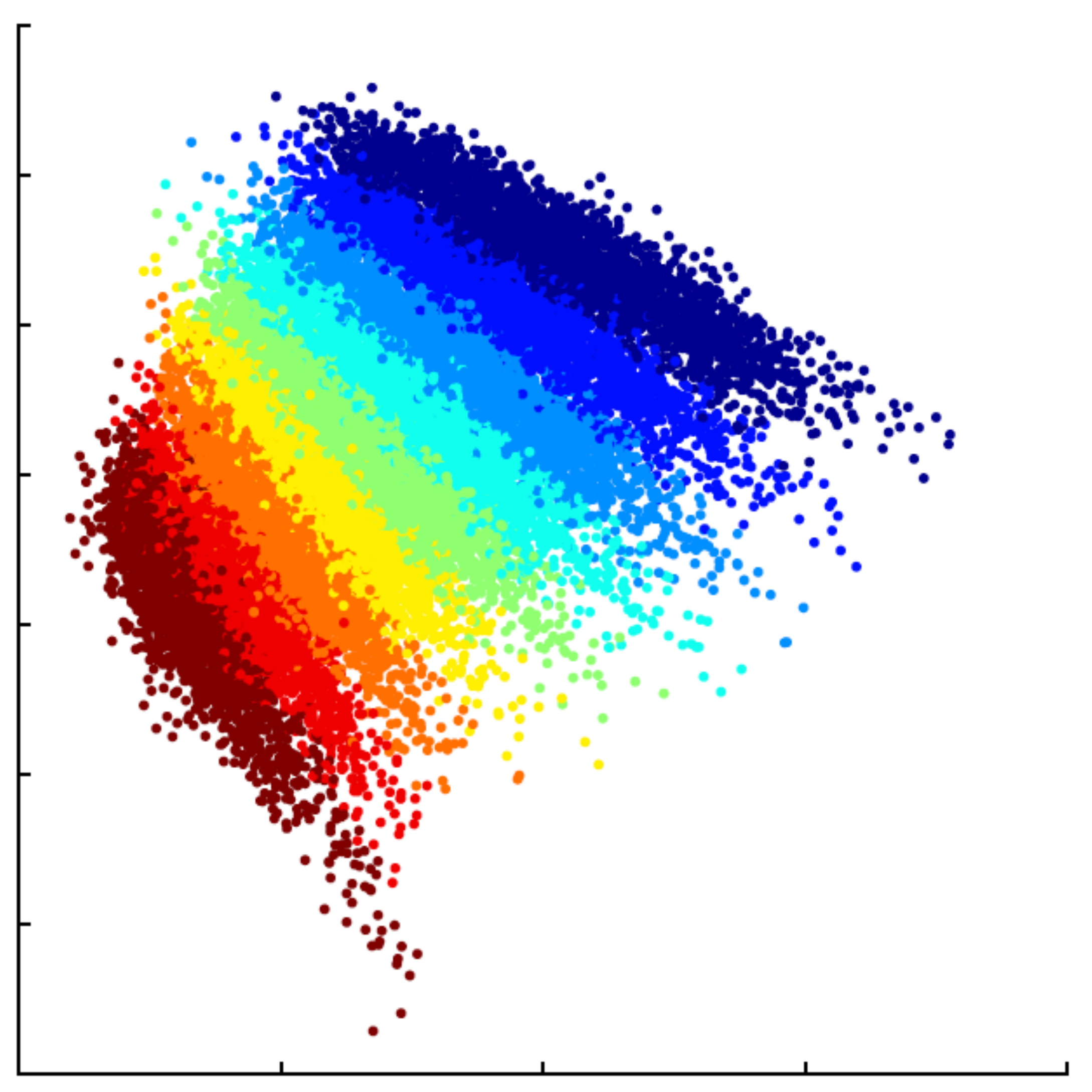}}\qquad%
  \subfloat{%
    \includegraphics[width=\norbfigsize]%
    {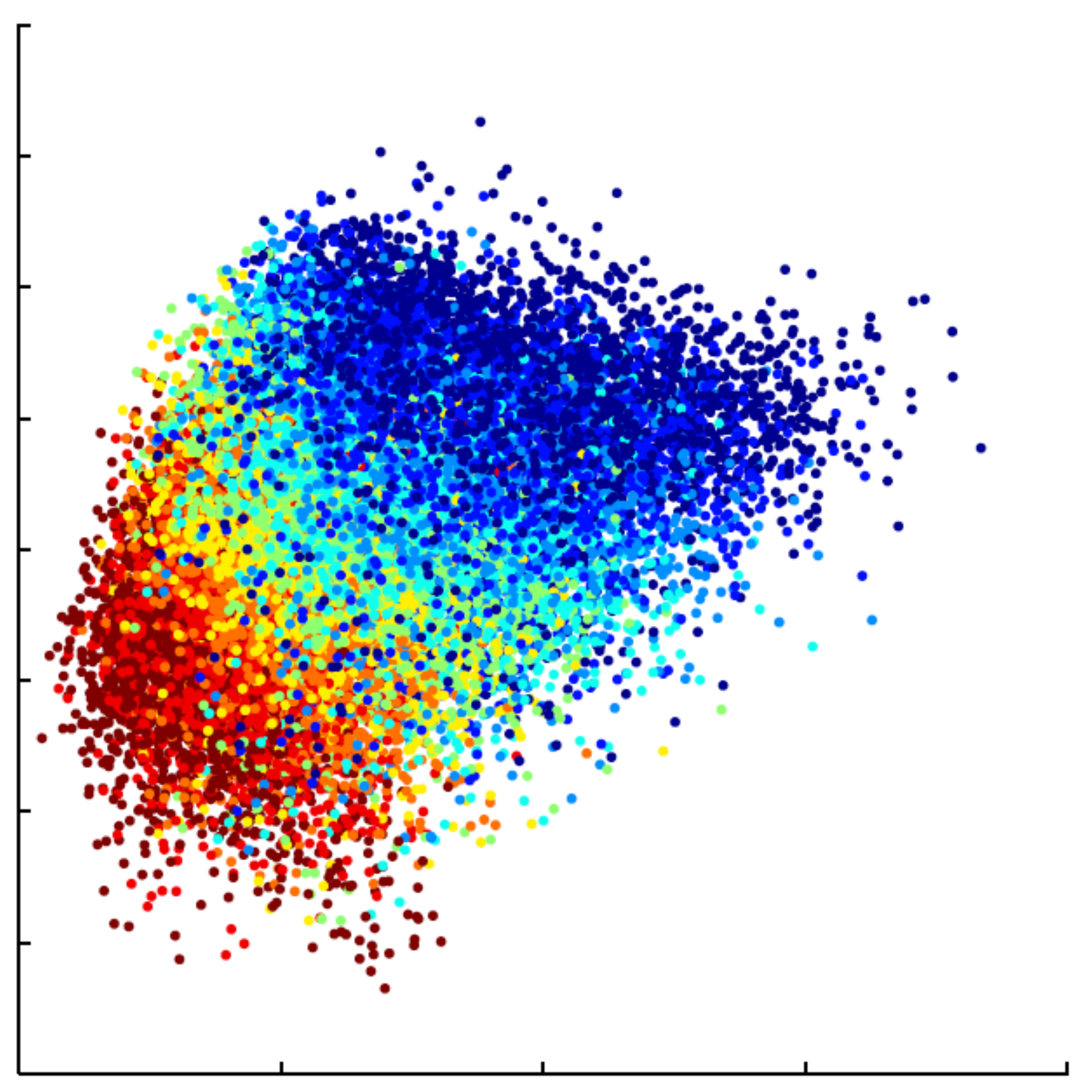}}\qquad%
  \\
  \vspace{-0.3cm}%
  \begin{sideways}~~~~~Lighting\end{sideways}
  \subfloat{%
    \includegraphics[width=\norbfigsize]%
    {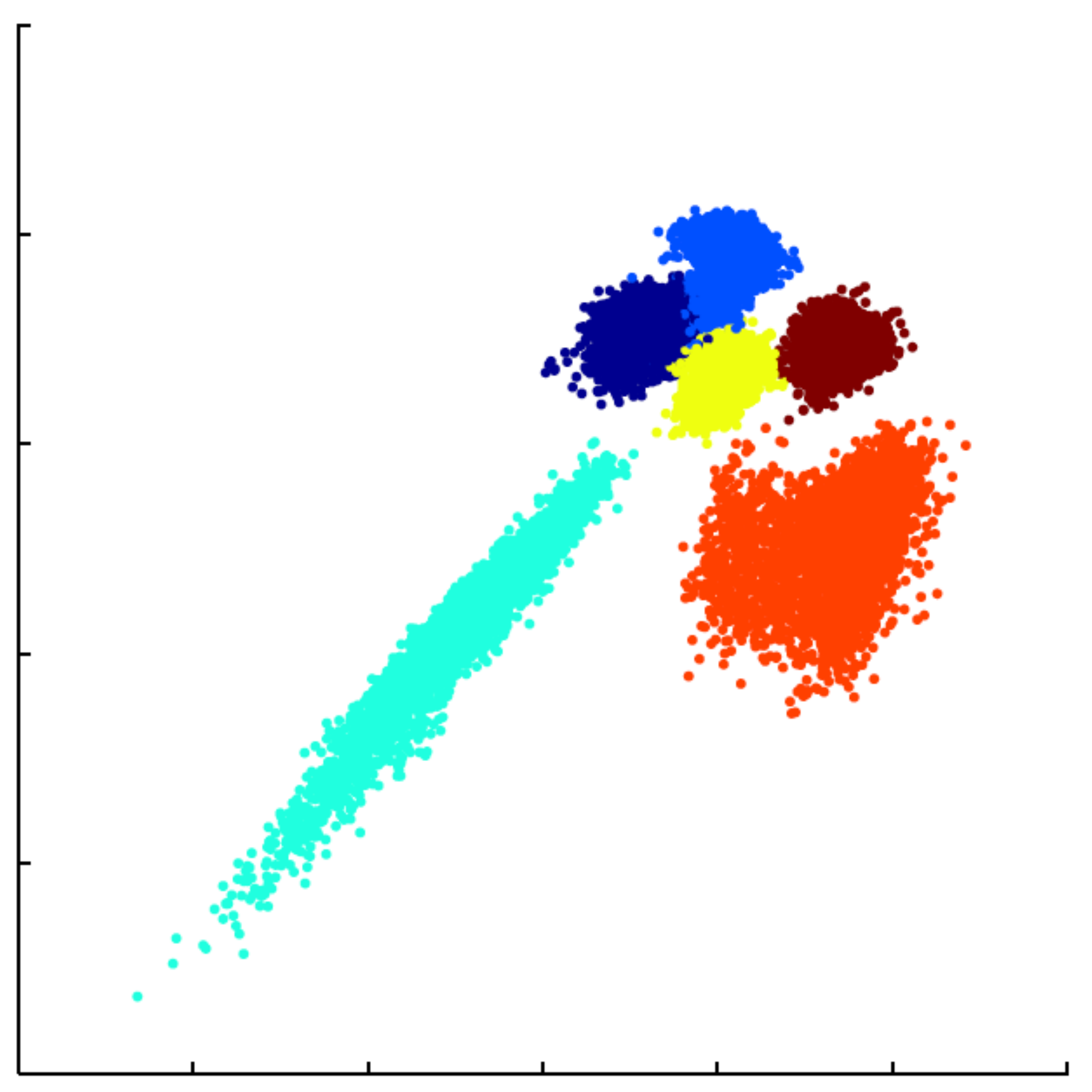}}\qquad%
  \subfloat{%
    \includegraphics[width=\norbfigsize]%
    {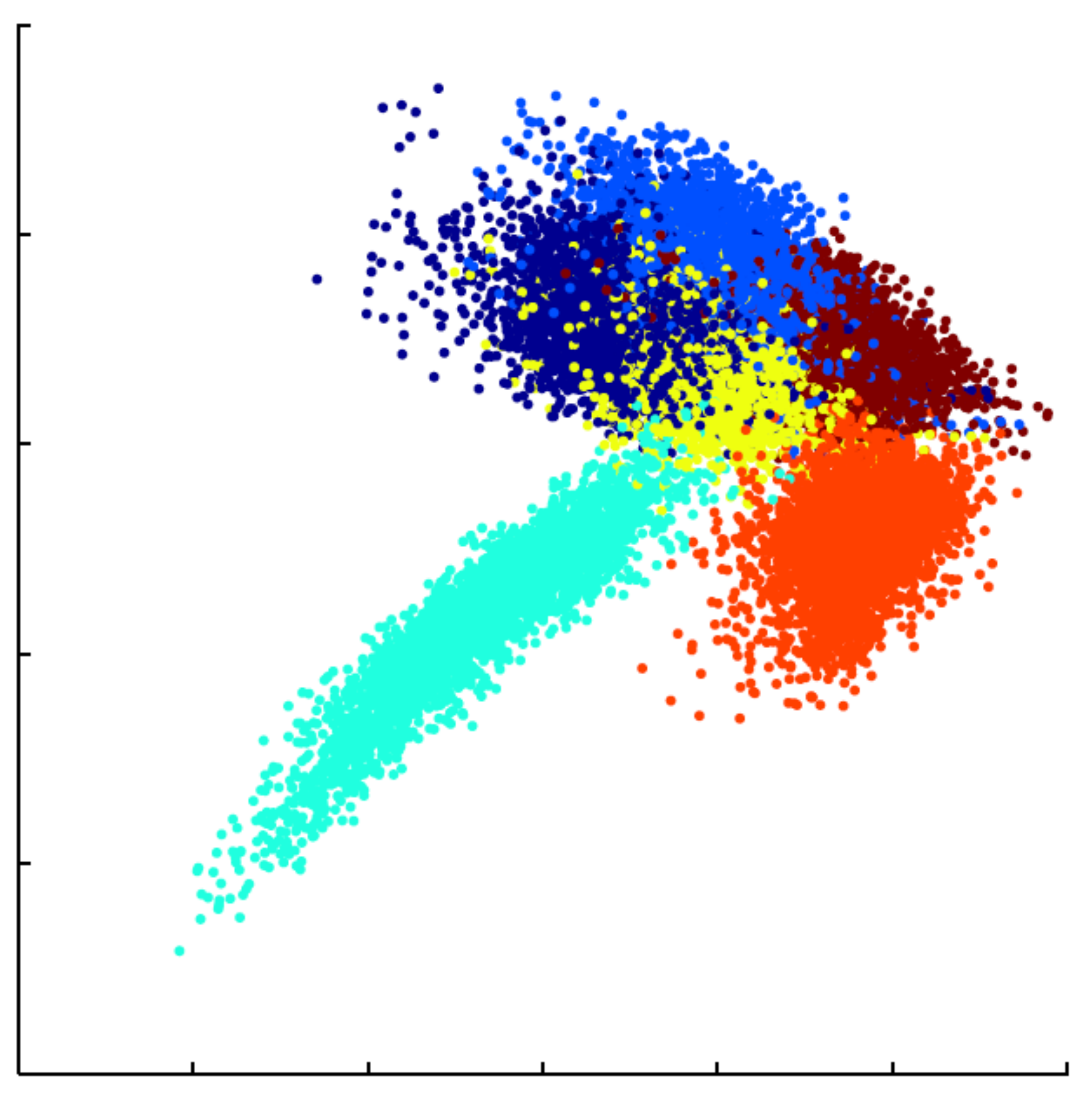}}\qquad%
  %\addtocounter{subfigure}{-3}
%\end{minipage}%
%\hfill%
 \caption{Visualisations of the NORB training (left) and
   test (right) data latent space representations in the NPGA,
   corresponding to class (first row), elevation (second row), and lighting (third row).
   Colors correspond to class labels.}%
 \label{fig:norb_visualisation}%
\vspace{-0.5cm}%
\end{figure}

\begin{figure}[t]
  \begin{tabular}[b]{p{0.32\textwidth}r}
    \toprule
    Model & Accuracy \\
    \midrule
    Autoencoder + 4(Log)reg ($\alpha=0.5$)& 85.97\% \\
    GPLVM & 88.44\% \\
    SGPLVM (4 GPs) & 89.02\% \\
    NPGA (4 GPs Lin -- $\alpha \medeq 0.5$) & 92.09\% \\  
    Autoencoder & 92.75\% \\
    Autoencoder + Logreg ($\alpha=0.5$)& 92.91\% \\
    NPGA (1 GP NN -- $\alpha \medeq 0.5$) & 93.03\% \\
    NPGA (1 GP Lin -- $\alpha \medeq 0.5$) & 93.12\% \\
    NPGA (4 GPs Mix -- $\alpha \medeq 0.5$) & 94.28\% \\
    \midrule
    K-Nearest Neighbors  & 83.4\% \\
    ~{\small \citep{lecun-etal-2004a}} & \\
    Gaussian SVM  & 88.4\% \\
    ~{\small \citep{salakhutdinov-larochelle-2010a}}&\\
    3 Layer DBN  & 91.69\% \\
    ~{\small \citep{salakhutdinov-larochelle-2010a}} & \\
    DBM: MF-FULL  & 92.77\% \\
    ~{\small \citep{salakhutdinov-larochelle-2010a}} & \\
    Third Order RBM & 93.5\% \\
    ~{\small \citep{nair-hinton-2009a}} & \\
   \bottomrule
 \end{tabular}
  \captionof{table}{Experimental results on the small NORB data test
    set. Relevant published results are shown for comparison.  NN, Lin
  and Mix indicate neural network, linear and a combination of neural
  network and periodic covariances respectively.}%
  \label{tab:norb_results}%
  \vskip -0.5cm%
\end{figure}

To address this question, an NPGA was employed with GPs mapping to
each of the four labels. Each GP was applied to a unique partition of
the hidden units of an autoencoder with 2400 NReLU units. A GP mapping
to the class labels was applied to half of the hidden units and
operated on a ${H\medeq 4}$ dimensional latent space.  The remaining
1200 units were divided evenly among GPs mapping to the three
auxiliary labels.  As the lighting labels are discrete, they were
treated similarly to the class labels, with~${H\medeq 2}$. The
elevation labels are continuous, so the GP was mapped directly to the
labels, with~${H\medeq 2}$. Finally, as the azimuth is a periodic
signal, a periodic kernel was used for the azimuth GP, with~${H\medeq
  1}$. This elucidates a major advantage of our approach, as the GP
provides flexibility that would be challenging with a parametric
mapping. 

This configuration was compared to an autoencoder
(${\alpha\medeq 0}$), an autoencoder with parametric logistic
regression guidance and a similar NPGA where only a GP to classes was
applied to all the hidden units. A back-constrained GPLVM and SGPLVM
were also applied to these data for comparison\footnote{The GPLVM and
  SGPLVM were applied to a 96 dimensional PCA of the data for
  computional tractability, used a neural net covariance mapping to
  the data, and otherwise used the same back-constraints, kernel
  configuration, and minibatch training as the NPGA.}. The
results\footnote{A validation set of 4300 training cases was withheld
  for parameter selection and early stopping.  Neural net covariances
  with fixed hyperparameters were used for each GP, except for the GP on
  the rotation label, which used a periodic kernel.  The raw pixels
  were corrupted by setting the value of 20\% of the pixels to zero
  for denoising autoencoder training. Each image was lighting and
  contrast normalized.  The error on the test set was evaluated using
  logistic regression on the hidden units of each model.}  are
reported in Table~\ref{tab:norb_results}.  A visualisation of the
structure learned by the GPs is shown in Figure
\ref{fig:norb_visualisation}.

The model with 4 GPs with nonlinear kernels obtains an accuracy of
94.28\% and significantly outperforms all other models, achieving to
our knowledge the best (non-convolutional) results for a shallow model
on this dataset.  Applying nonparametric guidance to all four of the
signals appears to separate the class relevant information from the
irrelevant transformations in the data.  Indeed, a logistic regression
classifier trained only on the 1200 hidden units on which the class GP
was applied achieves a test error of 94.02\%, implying that half of
the latent representation can be discarded with virtually no
discriminative penalty.

One interesting observation is that, for linear kernels, guidance with
respect to all labels decreases the performance compared to using
guidance only from the class label (from 93.03\% down to 92.09\%). An
autoencoder with parametric guidance to all four labels was tested as
well, mimicking the configuration of the NPGA, with two logistic and
two gaussian outputs operating on separate partitions of the hidden
units.  This model achieved only 86\% accuracy.  These observations
highlight the advantage of the GP formulation for supervised guidance,
which gives the flexibility of choosing an appropriate kernel for
different label mappings (e.g.\ a periodic kernel for the rotation
label).

\section{Conclusion}
In this paper we observe that the back-constrained GPLVM can be
interpreted as the infinite limit of a particular kind of autoencoder.
This relationship enables one to learn the encoder half of an
autoencoder while marginalizing over decoders.  We use this
theoretical connection to marginalize over functional mappings from
the latent space of the autoencoder to any auxiliary label
information.  The resulting nonparametric \textit{guidance} encourages
the autoencoder to encode a latent representation that captures
salient structure within the input data that is harmonious with the
labels.  Specifically, it enforces the requirement that a smooth
mapping exists from the hidden units to the auxiliary labels, without
choosing a particular parameterization.  By applying the approach to
two data sets, we show that the resulting nonparametrically guided
autoencoder improves the latent representation of an autoencoder with
respect to the discriminative task.  Finally, we demonstrate on the
NORB data that this model can also be used to \textit{discourage}
latent representations that capture statistical structure that is
known to be irrelevant through guiding the autoencoder to separate the
various sources of variation. This achieves state-of-the-art
performance for a shallow non-convolutional model on NORB.

\small

\bibliographystyle{abbrvnat}
\bibliography{draft}

\end{document}